\documentclass[10pt,twocolumn,letterpaper]{article}       

\usepackage{amssymb}
\usepackage{iccvcmr}
\usepackage{times}
\usepackage{epsfig}
\usepackage{graphicx}
\usepackage{amsmath}
\usepackage{amssymb}
\usepackage{booktabs}
\usepackage{multirow}
\usepackage{xcolor}
\usepackage{caption}
\usepackage{capt-of,etoolbox}

\usepackage[tight-spacing=true]{siunitx}

\usepackage[pagebackref=true,breaklinks=true,letterpaper=true,colorlinks,bookmarks=false]{hyperref}

\iccvfinalcopy 

\def\iccvPaperID{5778} 

\ificcvfinal\fi

\newcommand{\mycomment}[1]{}

\newcommand{\refsec}[1]{Sec.~\ref{#1}}
\newcommand{\reffig}[1]{Fig.~\ref{#1}}
\newcommand{\refeq}[1]{Eq.~\ref{#1}}

\newcommand{\ba}{\begin{eqnarray*}}
\newcommand{\ea}{\end{eqnarray*}}
\newcommand{\shp}{\mathcal{S}}
\newcommand{\sv}{s}
\newcommand{\shpbasis}{\mathbb{B}}

\newcommand{\matilde}{~}

\newcommand{\xx}{\mathbf{x}}
\newcommand{\prj}{\Pi}

\newcommand{\normone}[1]{\left\lVert#1\right\rVert_1}
\newcommand{\normtwo}[1]{\left\lVert#1\right\rVert_2}

\newcommand\blfootnote[1]{%
  \begingroup
  \renewcommand\thefootnote{}\footnote{#1}%
  \addtocounter{footnote}{-1}%
  \endgroup
}

\title{Lifting AutoEncoders: Unsupervised Learning of a Fully-Disentangled \\ 3D Morphable  Model using Deep Non-Rigid Structure from Motion}

\author{Mihir Sahasrabudhe*\\
CentraleSup\'{e}lec\\
\and
Zhixin Shu*\\
Stony Brook University\\
\and
Edward Bartrum\\
The Alan Turing Institute, UCL\\
\and
R\i{}za Alp G\"{u}ler\\
Ariel AI, Imperial College\\
\and
Dimitris Samaras\\
Stony Brook University\\
\and
Iasonas Kokkinos\\
Ariel AI, UCL\\
} 

\begin{document}

\twocolumn[{%
\renewcommand\twocolumn[1][]{#1}%
\maketitle
\begin{center}
    \centering
    \includegraphics[width=.999\textwidth]{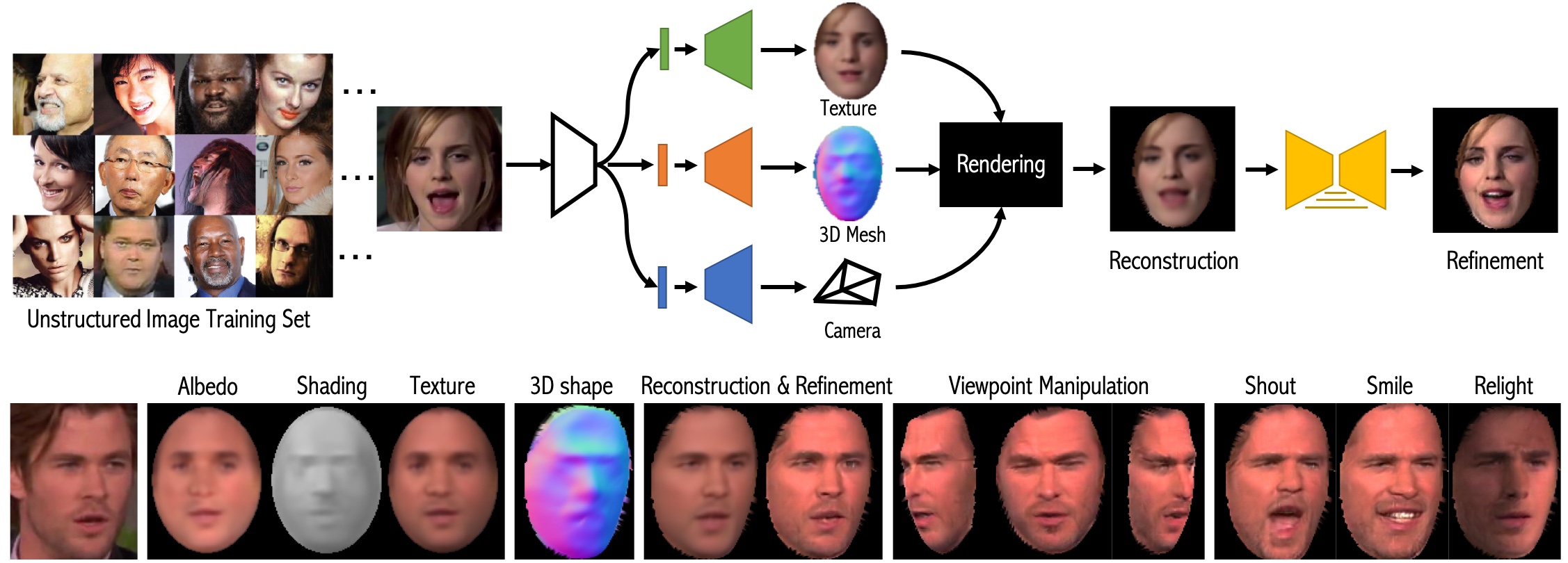}
    \captionof{figure}{We introduce Lifting AutoEncoders, a deep generative model of 3D shape variability that is learned from an unstructured photo collection without supervision. Having access to 3D allows us to disentangle the effects of viewpoint, non-rigid shape (due to identity/expression), illumination and albedo and perform  entirely controllable image synthesis.}
\end{center}%
}]

\blfootnote{* Indicating equal contributions.}

\begin{abstract}
In this work we introduce Lifting Autoencoders, a generative  3D surface-based model of object categories. We bring together ideas from non-rigid structure from motion,  image formation, and morphable models to learn a controllable, geometric model of 3D categories in an entirely unsupervised manner from an unstructured set of images. 
We exploit the 3D geometric nature of our model and use normal information to disentangle appearance into illumination, shading and albedo. We further use weak supervision to disentangle the non-rigid shape variability of human faces into identity and expression. 
We combine the 3D representation with a differentiable renderer to generate RGB images and  append an adversarially trained refinement network  to obtain sharp, photorealistic image reconstruction results. 
The learned generative model can be controlled  in terms of interpretable geometry and appearance factors, allowing us to perform photorealistic image manipulation of identity, expression, 3D  pose, and illumination properties. 
\end{abstract}

\section{Introduction}
Computer vision can be understood as the task of inverse graphics, namely the recovery of the scene that underlies an observed image. The scene factors that govern image formation primarily include surface geometry, camera position, material properties and illumination. These are independent of each other, but jointly determine the observed image intensities. 

In this work we incorporate these factors as disentangled variables in a deep generative model of an object category and tackle the problem of recovering all of them in an entirely unsupervised manner. We integrate in our network design ideas from classical computer vision, including structure-from-motion, spherical harmonic models of illumination and deformable models, and 
recover the three-dimensional geometry of a deformable object category in an entirely unsupervised manner from an unstructured collection of RGB images. We focus in particular on human faces and show that we can learn a three-dimensional morphable model of face geometry and appearance without access to any 3D training data, or manual labels.
We further show that by using weak supervision we can further disentangle identity and expression, leading to even more controllable 3D generative models. 

The resulting model allows us to generate photorealistic images of persons in a fully-controllable manner: we can manipulate 3D camera pose, expression, texture and illumination in terms of disentangled and interpretable low-dimensional variables. 

Our starting point is the Deforming AutoEncoder (DAE) model introduced in \cite{shu2018eccv} to learn an unsupervised deformable template model for an object category. DAEs incorporate deformations in the generative process of a deep autoencoder by associating pixels with the UV coordinates of a learned deformable template. As such, they disentangle appearance and shape variability and learn dense template-image correspondences in an unsupervised manner. 
 
We first introduce Lifting AutoEncoders (LAEs) to recover, and then exploit the underlying 3D geometry of an object category by interpreting the outputs of a DAE in terms of a 3D representation. For this we train a network task so as  minimize a Non-Rigid SfM minimization objective, which results is a low-dimensional morphable model of 3D shape, coupled with an estimate of the camera parameters. The resulting 3D reconstruction is coupled with a differentiable renderer \cite{kato2018renderer} that  propagates information from a 3D mesh to a 2D image, yielding a generative model for images that can be used  for both image reconstruction and manipulation.

Our second contribution consists in exploiting the 3D nature of our novel generative model to  further disentangle the image formation process. This is done in two complementary ways. For illumination modeling we use the 3D model to render normal maps and then shading images, which are combined with albedo maps to synthesize appearance. The resulting generative model  incorporates our spherical-harmonics-based \cite{zhang2005, wang2007, wang2009} modeling of  image formation, while still being end-to-end differentiable and controllable.  For shape modeling 
we use  sources of weak supervision to factor the shape variability into 3D pose, and non-rigid identity and expression, allowing us to control the expression or identity of a face by working with the appropriate latent variable code. 


\begin{figure*}
    \centering
    \includegraphics[width=.999\textwidth]{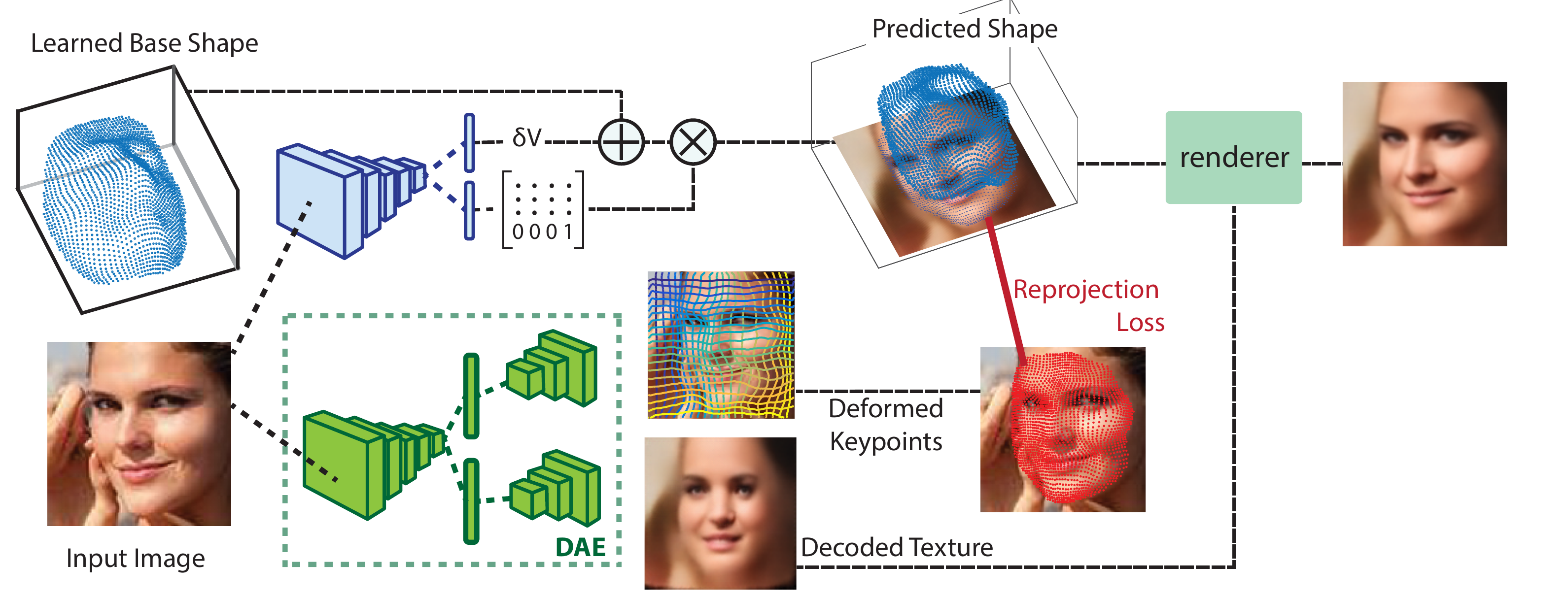}
    \captionof{figure}{Lifting AutoEncoders bring Non-Rigid Structure from Motion (NRSfM) into the problem of learning disentangled generative models for object categories. We start from a Deforming-AutoEncoder (DAE) that interprets images in terms of non-rigid, 2D warps between a template and an image. We train a Lifting AutoEncoder network by minimizing a NRSfM-based reprojection error between the learned, 3D Morphable Model-based vertices and their respective DAE-based positions. Combined with a differentiable renderer providing 3D-to-2D information. and an adversarially trained refinement network this provides us with an end-to-end trainable architecture for photorealistic image synthesis.}
\end{figure*}

Finally, we combine our reconstruction-driven architecture with an adversarially trained refinement network which allows us to generate photo-realistic images as its output. 

As a result of these advances we have a deep generative model that uses 3D geometry to model shape variability and provides us with a clearly disentangled representation of 3D shape in terms of identity, expression and camera pose  and appearance in terms of albedo and illumination/shading. We report quantitative results on a 3D landmark localization task and show multiple qualitative results of controllable photorealistic image generation.

\section{Previous work}





The task of disentangling deep models  can be understood as splitting the latent space of a network into independent sources of variation. In the case of learning generative models for computer vision, this amounts to uncovering the independent factors that contribute to image formation. This can both simplify learning, by injecting inductive biases about the data generation process, and can also lead to interpretable models that can controlled  by humans in terms of a limited number of degrees of freedom. This would for instance allow computer graphics to benefit from the advances in the learning of generative models. 

Over the past few years rapid progress has been made in the direction of  disentangling the latent space of deep models into dimensions that account for generic  factors of variation, such as identity and low-dimensional transformations \cite{infogan,brostow17,hinton10,WorrallGTB16,sundermeyer2018implicit}, or even non-rigid, dense deformations from appearance \cite{ZhouKAHE16,denseiccv17,ThewlisBV17a,shu2018eccv,wiles2018x2face}.
 Several of these techniques have made it into some of the most compelling photorealistic, controllable generative models of object categories \cite{pumarola2018ganimation,karras2018style}. 

Moving closer to graphics, recent works have aimed at exploiting our knowledge about image formation in generative modeling by replicating the inner workings of graphics engines in deep networks. 
On the synthesis side, geometry-driven generative models using intrinsic images \cite{ShuYHSSS17,DBLP:journals/corr/abs-1809-04696,sfsnetSengupta18} or 
the 2.5D image sketch \cite{DBLP:journals/corr/abs-1812-02725} as inputs to image synthesis networks have been shown to deliver sharper, more controllable image and video \cite{Kim:2018:DVP:3197517.3201283}
synthesis results. 
On the analysis side, several works have aimed at intrinsic image decomposition \cite{barrow1978recovering} using energy minimization, e.g \cite{Gehler,video}. The disentanglement of image formation into all of its constituent sources (surface  normals, illumination and albedo) was first pursued in \cite{Barron:EECS-2013-117}, where priors over the constituent variables were learned from generic scenes and then served as regularisers to complement the image reconstruction loss. More recently, deep learning-based works have aimed at learning the intrinsic image decomposition from synthetic supervision \cite{maire}, self supervision \cite{Janner0KYT17} or multi-view supervision \cite{smith}. 

These works can be understood in D. Marr's terms as getting 2.5D proxies to 3D geometry, which could eventually lead to 3D reconstruction \cite{marrnet}: texture is determined by shading, shading is  obtained from normals and illumination, and normals are obtained from the 3D geometry. This leads to the task of 3D geometry estimation as being the key to a thorough disentanglement of image formation. 



Despite these advances, the disentanglement of the three-dimensional world geometry from the remaining aspects of image formation still remains  very recent in deep learning. Effectively all works  addressing aspects related to 3D geometry rely on paired data for training, e.g. multiple views of the same object \cite{tulsiani2017multi}, videos \cite{novotny2017learning} or some pre-existing 3D mesh representation that is the starting point for further disentanglement \cite{Freeman18,sfsnet,yao20183d,tewari2018fml}
or self-supervision \cite{zhou2017unsupervised}.
This however leaves open the question of how one can learn about the three-dimensional world simply by observing a set of unstructured images. 

Very recently, a few works have started tackling the problem of recovering the three-dimensional geometry of objects from more limited information. 
In \cite{cmrKanazawa18} the authors used segmentation masks and keypoints to learn a CNN-driven 3D morphable model of birds, trained in tandem with a differentiable renderer module \cite{kato2018renderer}. Apart from the combination with an end-to-end learnable framework, this  requires however the same level of manual annotation (keypoints and masks) that earlier works had used to lift object categories to 3D \cite{carreira2016lifting}. A similar approach  has been proposed in \cite{nonlinear-3d-face-morphable-model} to learn morphable models from keypoint annotations. 

The LiftNet architecture proposed more recently by \cite{Wiles18b} uses a 3D geometry-based reprojection loss  to train a depth regression FCN by using  correspondences of object instances during training. This however is missing the surface-based representation of a given category, and is using geometry only implicitly, in its loss function - the network itself is a standard FCN.

The unsupervised training of volumetric CNNs was originally proposed in \cite{NIPS2016_6600} using toy examples and mostly binary masks. Most recently, a GAN-based volumetric model of object categories was introduced in
\cite{henzler2018escaping}, showing that one can recover 3D geometry from an unstructured photo collection using adversarial  training. Still, this is far from a rendering pipeline, in the sense that the effects of illumination and texture are coupled together, and the volumetric representation implies limitations in resolution. 

Even though these works  present exciting progress in the direction of deep 3D reconstruction, they fall short of providing us with a model that operates like a full-blown rendering pipeline. 
By contrast in our work we propose for the first time a deep learning-based method that recovers a three-dimensional, surface-based, deformable template of an object category from an unorganized set of images, leading to controllable photorealistic image synthesis. 

We do so by relying on on Non-Rigid Structure from Motion (NRSfM). Rigid SFM is a mature technology, with efficient  algorithms existing for multiple decades years \cite{tomasi1992shape,MVG2003},  systems for large-scale, city-level 3D reconstruction were introduced  a decade ago \cite{rome}, while high-performing systems are now publicly available \cite{colmap}.  Rigid SFM has very recently been
revisited from the deep learning viewpoint, leading to exciting new results
\cite{ummenhofer2017demon,zhou2017unsupervised}. 

In contrast, NRSfM is still a largely unsolved problem. 
Developed originally to establish a 3D model of a deformable object by observing its motion \cite{bregler2000recovering} it was developed to solve increasingly accurately the underlying mathematical optimization problems \cite{torresani2008nonrigid,paladini2009factorization,akhter2009nonrigid,dai2014simple}, extending to dense reconstruction
\cite{garg2013dense},  lifting object categories from keypoints and masks
\cite{carreira2016lifting,cmrKanazawa18}, incorporating spatio-temporal priors \cite{simon2014separable} and illumination models  
\cite{liu2017better}, while leading to impressively high-resolution 3D Reconstruction results
\cite{gotardo2015photogeometric,liu2017better,HernandezHCM17}.
In \cite{dnrsfm} it has recently been proposed to represent non-rigid variability in terms of a deep architecture - but still the work relies on given point correspondences between instances of the same category. By contrast, our proposed method has a simple, linear model for the shape variability, as classical morphable models, but establishes the correspondences automatically.

Earlier NRSfM-based work has shown that 3D morphable model learning is possible in particular for human faces \cite{kemelmacher2013internet,kemelmacher2012collection,kemelmacher2011face} by using a carefully designed, flow-based algorithm to uncover the organization of the image collection - effectively weaving a network of connections between pixels of  images, and feeding this into  NRSfM. As we now show this is no longer necessary - we delegate the task of establishing correspondences across image pixels of multiple images to a Deforming Auto-Encoder \cite{shu2018eccv} and proceed to lifting images through an end-to-end trainable deep network as we now describe. 
\label{nrsfmprior}. Several other works have shown that combining a prior template about the object category shape with video allows for an improved 3D reconstruction of the underlying geometry, both for faces \cite{ThiesZSTN16,DBLP:journals/corr/abs-1812-07603,roussos18} and quadrupeds
 \cite{DBLP:journals/corr/abs-1811-05804}. However, these methods still require multiple videos and a template, while our method does not. We intend to explore the use of video-based supervision in future work.

\section{Lifting AutoEncoders}

We start by briefly describing Deforming AutoEncoders, as these are the starting point of our work. 
We then turn to our novel contributions of 3D lifting in \refsec{lifting} and shape disentanglement in  \refsec{shape}.

\subsection{DAEs: from image collections to deformations}
\begin{figure*}
\includegraphics[width=.49\linewidth]{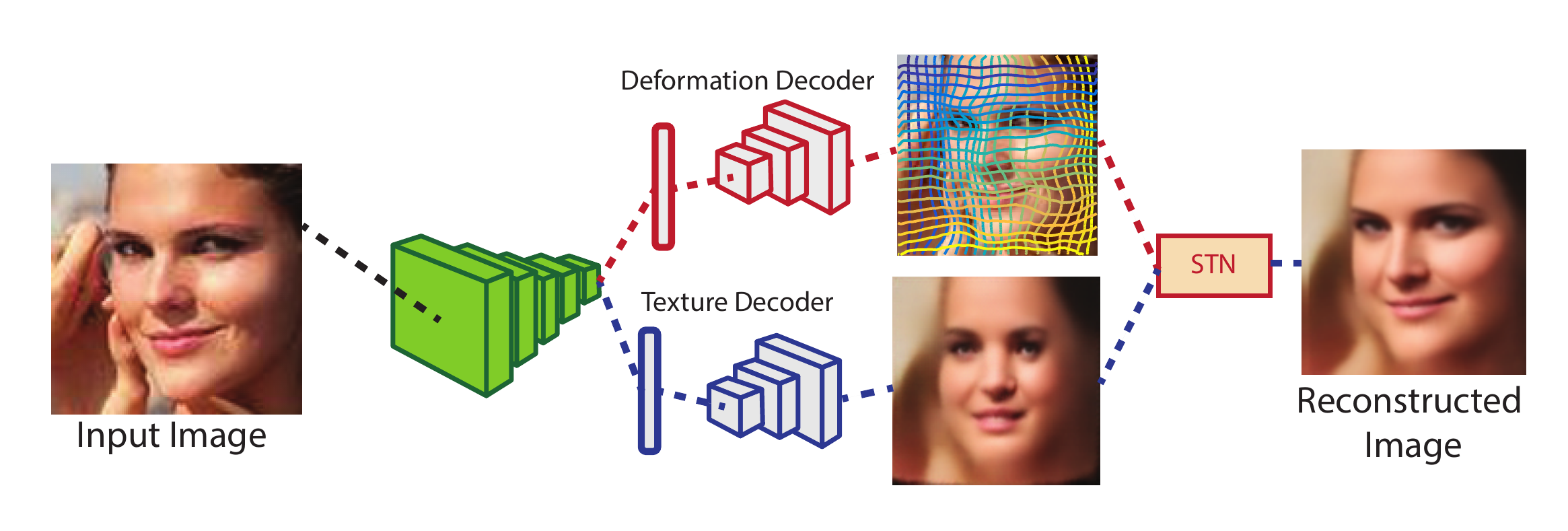}\hspace{.005\linewidth}
\includegraphics[width=.49\linewidth]{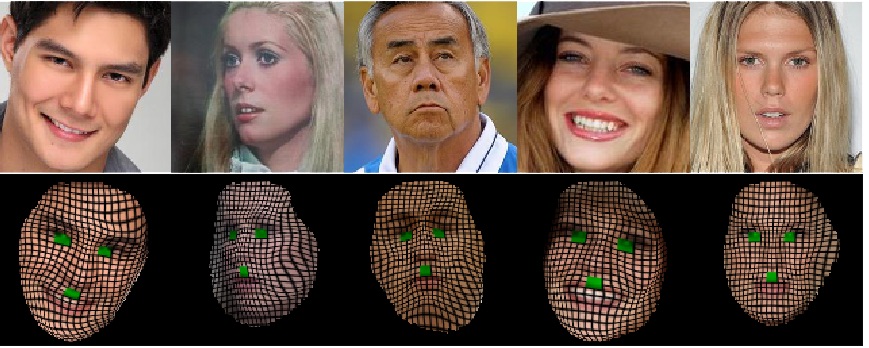} 

\caption{Deforming AutoEncoders (left) model image variability in terms of planar deformations of a deformation-free object template, that has to reconstruct the observed image when warped back to the image domain. As shown in the right part, learning this warping function establishes a common coordinate system for all images of the same category, effectively bringing into  correspondence pixels of objects with widely different appearance.}
\label{fig:dae}
\end{figure*}

     Deforming Autoencoders, introduced in \cite{shu2018eccv}, and shown in \reffig{fig:dae}, follow the deformable template paradigm and model image generation through a combination of appearance synthesis in a canonical coordinate system and a spatial deformation that warps the appearance (or, `texture') to the observed image coordinates. The resulting model  disentangles shape and appearance in an entirely unsupervised manner, while using solely an image reconstruction loss for training.
     Training a DAE is in principle an ill-posed problem, since the model could  learn to model shape variability in terms of appearance and recover a constant, identity deformation, resulting in a standard AutoEncoder. This is handled in practice by forcing the network to model shape variability through the deformation branch by reducing the dimensionality of the latent vector for textures. Further details for training DAEs are provided in \cite{shu2018eccv}.
     

\subsection{LAEs: 3D structure-from-deformations}
\label{lifting}
We now turn to the  problem of recovering the 3D geometry of an object category from an unstructured set of images. For this we rely on DAEs to identify corresponding points across this image set, and  address our problem by training a network to minimize an objective function that is inspired from Non-Rigid Structure from Motion (NRSfM). 
 Our central observation is that DAEs provide us with an image representation on which  NRSfM optimization objectives can be easily applied. 
In particular, disentangling appearance and deformation labels all image positions that correspond to a single template  point with a common, discovered $UV$ value.
LAEs take this a step further, and  interpret the DAE's UV decoding outputs as indicating the positions where an underlying 3D object surface position projects to the image plane. The task of an LAE is to then infer a 3D structure that can  successfully project to all of the observed 2D points. 

Given that we want to handle a deformable, non-rigid object category, we introduce a loss function that is inspired from Non-Rigid Structure from Motion, and optimize  with respect to it.
The variables involved in the optimization include (a) the statistical 3D shape representation, represented  in terms of a linear basis (b) the per-instance expansion coefficients on this basis and (c) the per-instance 3D camera parameters. We note that in standard NRSfM  all of the observations come from a common instance that is observed in time - by constrast in our case every training sample stems from a different instance of the same category, and it is only thanks to the DAE-based preprocessing that these distinct instances become commensurate.



%
\subsection{3D Lifting Objective}
Our 3D structure inference  task amounts to the recovery of  a surface model that maps an intrinsic coordinate space $(u,v)$ to 3D coordinates: $S(u,v) \to \mathbb{R}^3$. 
Even though the underlying model is continuous,  our implementation is discrete: we consider a set of 3D points sampled uniformly on a cartesian grid in intrinsic coordinates,
\begin{gather}
\shp_i = S(u_i,v_i), \matilde (u_i,v_i) \in D \times D, \\
\mathrm{with}\matilde D = \left\{0,\frac{1}{n},\frac{2}{n},\ldots,1\right\},\matilde i =1, \ldots, N=(n+1)^2,
\end{gather}
 where $n$ determines the spatial resolution at which we discretize the surface. We parameterize the three-dimensional position of these vertices in terms of a low-dimensional linear model, that captures the dominant modes of variation around a mean shape $\shpbasis^0$,
\begin{eqnarray}
\shp_i = \shpbasis^0 + \sum_{s=1}^S \sv_s \shpbasis^s_i. \label{eq:shape}
\end{eqnarray}
In morphable models \cite{vetter,Booth2018} the mean shape and deformation basis elements are learned by PCA on a set of aligned 3D shapes, but in our case we discover them from 2D by solving an NRSfM minimization problem that involves the projection to an unknown camera viewpoint.

In particular we consider scaled orthographic projection $\Pi$ through a camera described by a rotation matrix $R$ and translation vector $t$. Under this assumption, the 3D surface points project to the points $\xx_i$, given by
\begin{eqnarray}
\xx_i &=& \prj  \left[ R \shp_i\right] + t, \quad
\prj = \left[\begin{array}{ccc} \sigma &  0 & 0 \\  0 & \sigma & 0 \end{array}\right] \label{eq:proj},
\end{eqnarray}
where $\sigma$ defines a global scaling.

We measure the quality of a 3D reconstruction in terms of the Euclidean distance of the predicted projection of a 3D point and its actual position in the image. In our case a 3D point $\shp_i$ is associated with surface coordinate $(u_i,v_i)$, we therefore penalize its distance from the image position $\hat{x}_i$ that the DAE's deformation decoder labels as $u_i,v_i$:
\begin{eqnarray}
\hat{\xx}_i = \hat{\xx}: \mathrm{arg}\mathrm{min}_{\xx} \|DAE(\xx) - (u_i,v_i)\|_2
\end{eqnarray}
In practice we return the image point $\xx$ where the DAE's prediction is closest to $(u_i,v_i)$; if no point is sufficiently close we declare that point $i$ is missing, setting a visibility variable $\nu_i$ to zero. We treat $\hat{\xx}$ and  $\nu$ as data terms, which specify the constraints that our learned 3D model must meet: the 3D points $\shp_i$ must project to points $\xx_i$ that lie close to their visible 2D counterparts, $\hat{\xx}_i$. We express this \emph{reprojection objective} in terms of the remaining variables:
\begin{eqnarray}
L(R,t,\sigma,\sv,\shpbasis) = 
\sum_{i=1}^N \nu_i \|\hat{\xx}_i - \xx_i(R,t,\sigma,\shp,\sv)\|_2 \label{eq:reprojection}
\end{eqnarray}
where we have expressed $\xx_i$ as a differentiable function of $R,t,\shp,\sv$ through \refeq{eq:proj} and \refeq{eq:shape}.

For a set of $K$ images we have different camera and shape parameters $(R_k,t_k),\sv_k, k=1,\ldots,K$  since we consider a non-rigid object seen from different viewpoints. The basis elements $\shpbasis$  are however considered to be common across all images, since they describe the inherent shape variability of the whole category. Our 3D non-rigid reconstruction problem  thus becomes:
\begin{eqnarray}
\mathcal{L}_{3D} = \sum_{k=1}^{K} L(R^k,t^k,\sigma^k,\sv^k,\shpbasis) \label{eq:sfm}
\end{eqnarray}

\subsection{LAE learning via Deep NRSfM}
Minimizing the objective of \refeq{eq:sfm} amounts to the common Non-Rigid Structure-from-Motion objective \cite{bregler2000recovering,torresani2008nonrigid,paladini2009factorization,akhter2009nonrigid,dai2014simple}. 
 Even though highly efficient and scalable algorithms have been proposed for its minimization, we would only consider them for initialization, since  we want 3D Lifting to be a component of a larger deep generative model of images. We do not use any such technique, in order to simplify our model's training, implementing it as a single deep network training process. 
 
The approach we take is to handle the shape basis $\shpbasis$ as the parameters of a linear `morphable' layer, tasked with learning the shape model for our object category.
We train this layer in tandem with complementary, multi-layer network branches that regress from the image to (a) the expansion  coefficients $\sv^k$, (b) the Euler angles/rotation matrix $R^k$, and (c) the displacement vector  $t^k$ describing the camera position. 
In the limit of very large hidden vectors the related angle/displacement/coefficient heads could simply memorize the optimal values per image, as dictated by the optimization of \refeq{eq:sfm}. With a  smaller number of hidden units these heads learn to successfully regress camera and shape vectors and can generalize to unseen images. As such, they are  components of a larger deep network that can learn to reconstruct an image in 3D - a task we refer to as Deep NRSfM.

If we only train a network to optimize this objective we obtain a network that can interpret a given image in terms of its 3D geometry, as expressed by the 3D camera position (rigid pose) and the instance-specific expansion coefficients (non-rigid shape). Having established this, we can conclude the task of image synthesis by projecting the 3D surface back to 2D. 
For this we combine the 3D lifting network  with a differentiable renderer \cite{kato2018renderer}, and bring the synthesized texture image in correspondence with the image coordinates. The resulting network is an end-to-end trainable pipeline for image generation that passes through a full-blown, 3D reconstruction process. 

Having established a controllable, 3D-based rendering pipeline, we turn to photorealistic synthesis. For this we further refine the rendered image by a U-Net \cite{ronneberger2015u} architecture that takes as input the reconstructed image and augments the visual plausibility. This refinement module is trained using two losses, firstly an $L_2$ loss to reconstruct the input image and secondly an adversarial loss to provide photorealism. 
The results of this module are demonstrated in Figure \ref{fig:refinement} - we see that while keeping intact the image generation process, we achieve a substantially more realistic synthesis.  







\section{Geometry-Based Disentanglement}

A Lifting AutoEncoder provides us with a disentangled representation of images in terms of 3D rotation, non-rigid deformation, and texture, leading to controllable image synthesis.

In this section we show that having access to the underlying 3D scene behind an image allows to further decompose the image generation into distinct, controllable sub-models, in the same way that one would do within a graphics engine. These contributions rely on  certain assumptions and data that are reasonable for human faces, but could also apply to several other categories.

We first describe in \refsec{appearance} how surface-based normal estimation allows us to disentangle appearance into albedo and shading using a physics-based model of illumination. In \refsec{shape} we then turn to learning a more fine-grained model of 3D shape and use weak supervision  to disentangle per-instance non-rigid shape into expression and identity.

\subsection{LAE-lux: Disentangling Shading and Albedo}
\label{appearance}
Given the 3D reconstruction of a face we can use certain assumptions about image formation that lead to physically-plausible illumination modeling. For this
we extend LAE with albedo-shading disentangling, giving rise to LAE-lux where we explicitly model illumination. 

As in several recent works \cite{ShuYHSSS17,sfsnetSengupta18} we consider a Lambertian reflectance model for human faces and adopt the  Spherical Harmonic model to model the effects of illumination on appearance \cite{zhang2005, wang2007, wang2009}. 
We pursue the intrinsic decomposition ~\cite{barrow1978recovering} of the canonical texture $T$ into albedo, $A$ and shading, $S$: 
\begin{equation}
    T = S \odot A
 \label{eq:intrinsic}
\end{equation}
where $\odot$ denotes Hadamard product, by constraining the shading image to be connected to the normals delivered by the LAE surface. 

In particular, denoting by $L$ the representation of the scene-specific spherical harmonic illumination vector, and by $H(N(x))$ the representation of the local normal field $N(x)$ on the first 9 spherical harmonic coefficients, we consider that the local shading, $S(x)$ is expressed as an inner product:
\begin{equation}
S(x) = \langle L,H(N(x))\rangle.
\end{equation}
As such the shading field can be obtained by a linear layer that is driven by regressed illumination coefficients $L$ and the surface-based harmonic field, $H(N(x))$.
Given $S(x)$, the texture can then be obtained from albedo  and shading images according to Eq.~\ref{eq:intrinsic}.

In practice, the normal field we estimate is not as detailed as would be needed, e.g. to capture sharp corners, while the illumination coefficients can be inaccurate. To compensate for this, we first render an estimate of the shading $S^{\text{render}}$ with spherical harmonics parameters $L$ and normal maps $N$and then use a U-Net to refine it, obtaining $S^\text{adapted}$.


\begin{figure}[]
 \centering
 \includegraphics[width=0.99\linewidth]{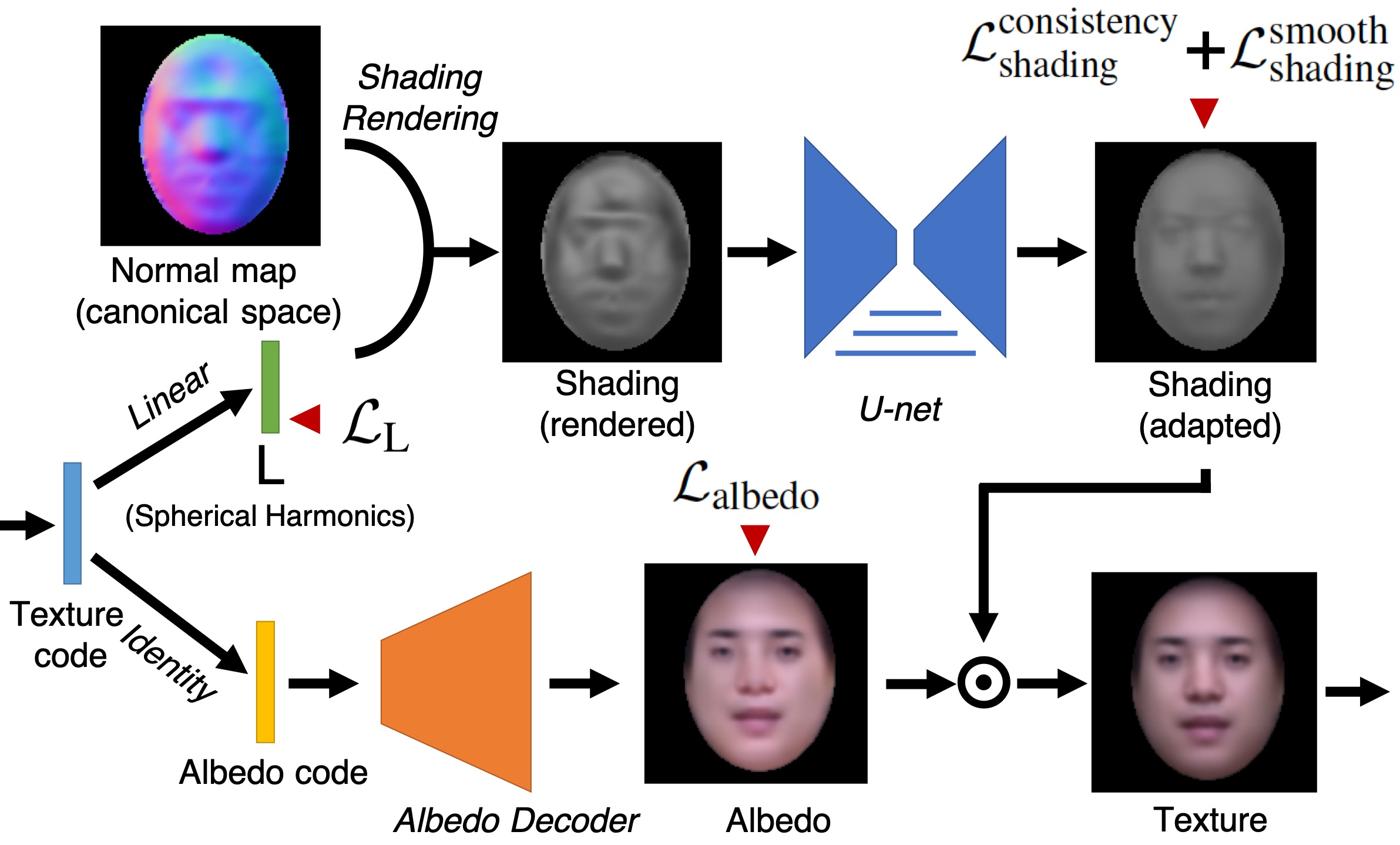}
 \caption{Texture decoder for LAE-lux: disentangling albedo and illumination with 3D shape and Spherical Harmonics representation for illumination.}
 \label{fig:albedonet}
\end{figure}








In our experiments we have initialized LAE-lux with a converged LAE, discarded the last layer of the LAE's texture prediction and replaced it with the inrinsic predictor outlined above. The albedo image is obtained through an albedo decoder that has an identical architecture to the texture decoder in DAE. 
The latent code for albedo $Z_A$ and the spherical harmonics  parameters $L$ are obtained as separate linear layers that process the penultimate layer of the texture encoder. 

In training, only the texture decoders are updated while other encoding and decoding networks are fixed. When instead  training everything jointly from scratch  we  observed implausible disentanglement results, presumably due to the ill-posed nature of the decomposition problem.

Given that the shading-albedo decomposition is an ill-posed problem, we further use a combination of losses that 
capture increasingly detailed prior knowledge about the desired solution. 
First, as in \cite{ShuYHSSS17} we employ intrinsic image-based smoothness losses on albedo and shading:
\begin{equation}
\mathcal{L}_\text{shading}^\text{smooth} = \lambda_{\text{shade}}\normtwo{ \nabla S^{\text{adapted}}},
\mathcal{L}_\text{albedo} = \lambda_{\text{albedo}} \normone{ \nabla A},
\end{equation}
where $\nabla$ represents the spatial gradient, which means that we allow the albedo to have sharp discontinuities, while the shading image should have mostly smooth variations \cite{samaras2000}. In our experiment, we set $\lambda_{\text{shade}} = \num{1e-4} $ and $\lambda_{\text{albedo}}  = \num{2e-6}$.

Second, we compute a deterministic estimate $\hat{L}$ of the illumination parameters and penalize its distance to the regressed illumination values: 
\begin{equation}
\mathcal{L}_\text{L} = \normtwo{L-\hat{L}}
\label{eq:sphericalloss}.
\end{equation}
More specifically, $\hat{L}$ is based on the crude assumption that the face's albedo is constant, $\hat{A}(x) =0.5$, where we treat albedo as a grayscale. 
Even though clearly very rough, this assumption captures the fact that a face is largely uniform, and allows us to compute a proxy to the shading in terms of 
$\hat{S} = T \oslash \hat{A}$ where $\oslash$ denotes Hadamard division.
We subsequently compute the approximation $\hat{L}$ from $\hat{S}$ and the harmonic field $H(N)$ using  least squares. For face images, similar to~\cite{ShuYHSSS17}, ${\hat{L}}$ serves as a reasonably rough approximation of the illumination coefficient and is used for weak supervision in \refeq{eq:sphericalloss}.

Finally, the shading consistency loss regularizes the U-Net, and is designed to encourage the U-Net based adapted shading $S^{\text{adapted}}$ to be consistent with the shading rendered from the spherical harmonics representation $S^{\text{rendered}}$---
\begin{equation}
\mathcal{L}_\text{shading}^\text{consistency} = {\text{Huber}}(S^{\text{adapted}},S^{\text{rendered}}),
\label{eq:shadingconsistencyloss}
\end{equation}
where we use Huber loss for a robust regression since $S^{\text{rendered}}$ can contain some outlier pixels due to an imperfect 3D shape.





\subsection{Disentangling Expression, Identity and Pose}
\label{shape}

Having outlined our geometry-driven model for disentangling appearance variability into shading and albedo, we now turn to the task of disentangling the sources of shape variability.

In particular, we consider that face shape, as observed in an image is the composite effect of camera pose, identity and expression. Without some guidance the parameters controlling shape can be mixed - for instance accounting for the effects of camera rotation through non-rigid deformations of the face. 

We start by allowing our representation to separately model identity and expression, and then turn to forcing it to disentangle pose, identity and expression. 

For a given identity we can understand expression-based shape variability in terms of deviation from a neutral pose.
We can consider that a reasonable approximation to this consists in using a separate linear basis $\shpbasis^{I}$ for identity and another for expression $\shpbasis^{E}$, which amounts to following  model:
\begin{eqnarray} 
\shp_i(\sv^I,\sv^E) = \shpbasis^0_i + \sum_{s=1}^{I} \sv^I_s \shpbasis^{I,s}_i
+  \sum_{s=1}^{E} \sv^E_s \shpbasis^{E,s}_i
\label{eq:shape_id_exp}
\end{eqnarray}

Even though the model is still linear and is at first sight equivalent, clearly separating the two subspaces means that we can control them through 
side information.
For instance when watching a video of a single person, or a single person from multiple viewpoints one can enforce  the identity expansion coefficients $\sv^I$ to remain constant through a siamese loss \cite{koch2015siamese}. This would force the training to model all of the person-specific variability through the remaining subspace, by changing the respective coefficients $\sv^E$ per image. 

Here we use the MultiPIE\cite{multipie2010} dataset to help disentangle the latent representation of person identity, facial expression, and pose (camera). MultiPIE is captured under a controlled environment and contains image pairs acquired under identical conditions with differences only in (1) facial expression, (2) camera position, and (3) illumination conditions. We use this dataset to disentangle the latent representation for shape into 
distinct components.

\newcommand{\SV}{S}
We denote by $\SV$ the concatenation of all shape parameters: $\SV = \left[\sv^C, \sv^I, \sv^E \right]$ and turn to the task of forcing the different components of $\SV$ to behave  as expected.
 We use facial expression distentangling as an example, and follow a similar procedure for pose and camera 
 disentangling. 
 Given an image $I_\text{exp}$ with known expression $\text{exp}$, we sample two more images. The first, $I_\text{exp}^{+}$ has the same facial expression  but different identity, pose, and illumination conditions. The second, $I_\text{exp}^{-}$, has a different facial expression but the same identity, pose and illumination condition as $I_\text{exp}$. We use 
 siamese training to encourage $I_\text{exp}$ and $I_\text{exp}^{+}$ to have similar latent representations for facial expression, and a triplet loss to ensure that $I_\text{exp}$ and $I_\text{exp}^{+}$ are closer in expression space than
 $I_\text{exp}$ and $I_\text{exp}^{-}$:
   \begin{eqnarray}
       \mathcal{L}_\text{expression} &=& \mathcal{L}_\text{expression}^{\text{similarity}} + \mathcal{L}_\text{expression}^{\text{triplet}},
   \text{where}\\
       \mathcal{L}_\text{expression}^{\text{similarity}} &=& \normtwo{f_{\text{exp}}(I_\text{exp}) - f_{\text{exp}}(I_\text{exp}^{+})},\\
   \mathcal{L}_\text{expression}^{\text{triplet}} &=& \max(0, 1+\normtwo{f_{\text{exp}}(I_\text{exp}) - f_{\text{exp}}(I_\text{exp}^{+})} \nonumber\\
    && - \normtwo{f_{\text{exp}}(I_\text{exp}) - f_{\text{exp}}(I_\text{exp}^{-})}).
   \end{eqnarray}     

Following a similar collection of triplets for the remaining sources of variability, we disentangle the latent code for shape in terms of camera pose, identity, and expression. With MultiPIE, the overall disentanglement objective for shape is hence
   \begin{equation}
       \mathcal{L}_{\text{disentangle}} = \mathcal{L}_\text{expression} + \mathcal{L}_\text{identity} + \mathcal{L}_\text{pose}, 
   \end{equation}
   where $\mathcal{L}_\text{identity}$ and $\mathcal{L}_\text{pose}$ are defined similarly to $\mathcal{L}_\text{expression}$. In our experiments, we used the scaling parameter for this loss, $\lambda_\text{disentangle} = 1$.

  \subsection{Complete Objective}
  Having introduced the losses that we use for disentangling, we now turn to forming our complete training objective. 
  
  We control the model learning with a regularization loss defined as follows:
   \begin{equation}
       \mathcal{L}_{\text{reg}} = \lambda_\text{scale} \sum_{k=1}^K \normtwo{\sigma_k}
              + \lambda_{\text{shape}}
              \sum_{k=1}^K \normtwo{\sum_{s=1}^S \sv^k_s \shpbasis^s},
  \end{equation}
   where $\sigma$ is the scaling parameter in \refeq{eq:proj} and $\sum_{s=1}^S \sv_s \shpbasis^s_i$ is the non-rigid deviation from the mean shape, $\shpbasis^0$. We use $\lambda_\text{scale}=0.01$, and $\lambda_{\text{shape}} = 0.1$ in all our experiments.    
   
  Combining this with the reprojection loss, $\mathcal{L}_\text{3D}$, defined in \refeq{eq:sfm}, 
   we can write the complete objective function, which is trained end-to-end:
   \begin{align}
      \begin{split}
       \mathcal{L}_{\text{total}} =~&\lambda_{\text{3D}} \cdot \mathcal{L}_{\text{3D}} ~+ \\
                                            &\lambda_\text{disentangle} \cdot \mathcal{L}_\text{disentangle} ~+ \\
                                            &\lambda_\text{scale} \cdot \mathcal{L}_\text{scale} ~+ \\
                                            &\lambda_{\text{shape}} \cdot \mathcal{L}_{\text{shape}}.
     \end{split}
   \end{align}
     In our experiments, we used the scaling factor for the 3D reprojection loss, $\lambda_{\text{3D}} = 50$. This relatively high scaling factor was chosen so that the reprojection loss is not overpowered by other losses at later training iterations. 
     
   For training the LAE-Lux, we also add the albedo-shading disentanglement losses, summarised by
   \begin{equation}
       \mathcal{L}_\text{lux} = \mathcal{L}_\text{shading}^\text{smooth} +
                                \mathcal{L}_\text{shading}^\text{consistensy} + 
                                \mathcal{L}_\text{albedo} +
                                \mathcal{L}_\text{L}.
   \end{equation}
   \mycomment{
   With MultiPIE, the overall disentanglement objective is hence
   \begin{equation}
       \mathcal{L}_{\text{disentangle}} = \mathcal{L}_\text{expression} + \mathcal{L}_\text{identity} + \mathcal{L}_\text{pose}.
   \end{equation}
   In our experiments, we used the scaling parameter for this loss, $\lambda_\text{disentangle} = 1$.
   }


\mycomment{
\section{Improving numbers and  quality}
Zhixin: Integrate Zhixin's DAE

Ed: Equivariance for UV (a-la Vedaldi)

Mihir: Validate a proper image encoder (e.g. resnet)

Mihir: Dataloaders for this dataset:
https://github.com/NVlabs/ffhq-dataset 

\section{Evaluation}

Mihir: Re-vive the 3D supervision-based results.

Ed: \href{http://www.robots.ox.ac.uk/~vgg/publications/2018/Wiles18a/wiles18a.pdf}{Link to Wiles et al.}

For Later: Explore semi-supervised training; 1K images with 3D supervision + 70K images without; do we better than with 1K with 3D supervision?https://www.overleaf.com/project/5c507ed3584d9620ef2e551c

\href{http://papers.nips.cc/paper/7657-unsupervised-learning-of-object-landmarks-through-conditional-image-generation.pdf}{Link to Jacab et al.}

\section{TODO:}
Iasonas: NN code
\begin{figure}
    \centering
    \includegraphics{results_vedaldi.png}
    \caption{results from Jacab et al.}
    \label{fig:my_label}
\end{figure}
}

 \section{Experiments}
 
   \begin{figure}[ht!]
       \centering
       \includegraphics[width=\linewidth]{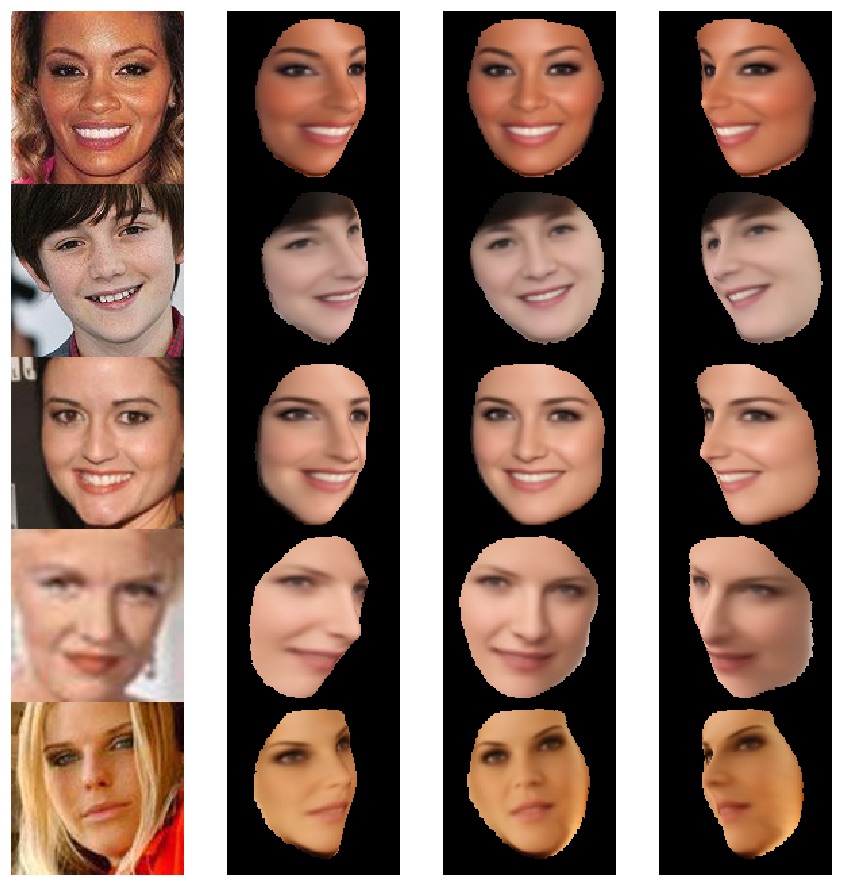}
       \caption{Visualizations of the learned 3D shapes from various yaw angles. Our reconstructions respect prominent face features, such as the nose, forehead and checks, allowing us to rotate an object reconstruction in 3D.}
       \label{fig:rotation}
   \end{figure}

  \begin{figure}[ht!]
   \centering
   \includegraphics[width=\linewidth]{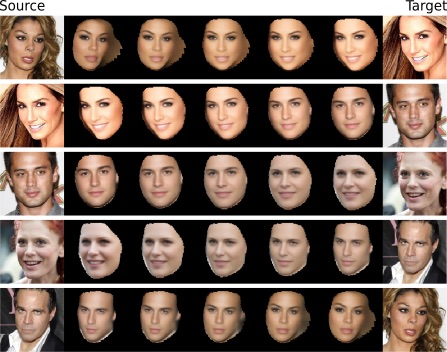}
   \caption{Interpolation on the shape, pose, and texture latent vectors. We show renderings of intermediate 3D shapes, with intermediate poses and textures, as we move around on all three latent spaces. }
   \label{fig:all_interpolation}
  \end{figure}
  
    \begin{figure}[ht!]
     \centering
     \includegraphics[width=\linewidth]{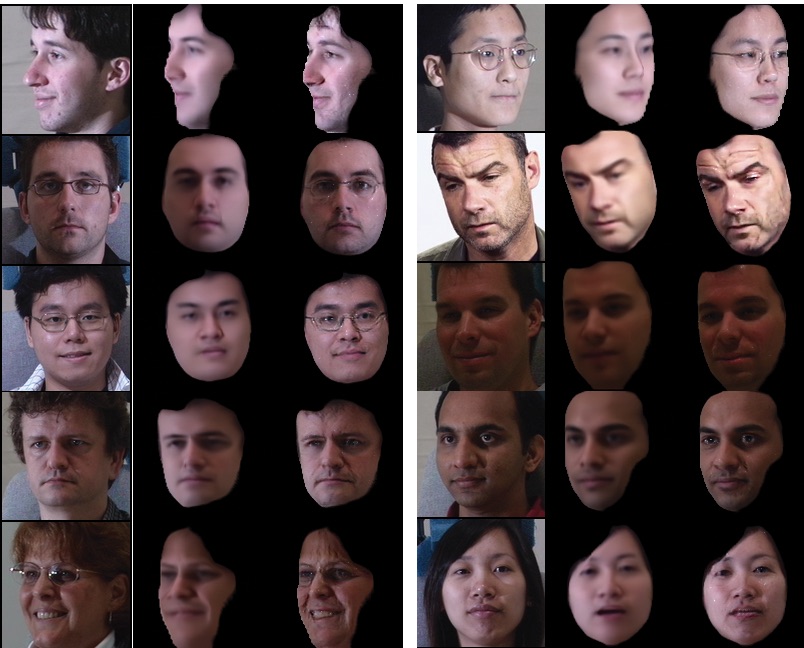}
     \caption{Photorealistic refinement: starting from an image reconstruction by an LAE (left), an adversarially-trained refinement network  adds details to increase the photorealism of a face (right). }
     \label{fig:refinement}
    \end{figure}
    
        \begin{figure}[ht!]
     \centering
     \includegraphics[width=1.\linewidth]{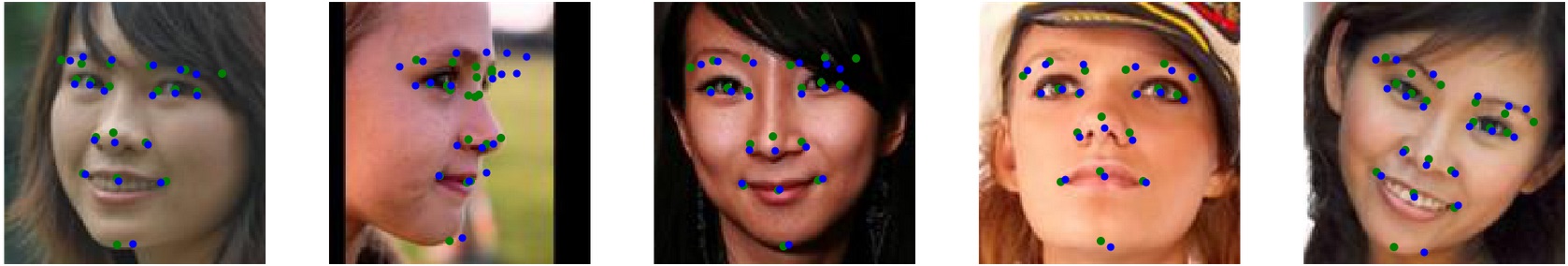}
     \caption{Landmark localization on a few AFLW2000 test images. We manually annotated landmarks in the UV space
     and visualized them after reprojection of the vertices. The LAE is able to localize landmarks for small 
     pose variability. Ground-truth landmarks are shown in green, whereas the predicted
    ones are shown in blue.}
     \label{fig:aflw_landmarks}
    \end{figure}
  \subsection{Architectural Choices}
   Our encoder and decoder architectures are similar to the ones employed in \cite{shu2018eccv}, but working on images
   of size $128 \times 128$ pixels instead of $64 \times 64$. We use convolutional neural networks with five stridedConv-batchNorm-leakyReLU layers in image encoders, which regress the expansion coefficients $\sv$s.
   Image decoders consist similarly of five stridedDeconv-batchNorm-ReLU layers. 
   
  In all of these experiments the training process was started with a base learning rate of $0.0001$, which was reduced by a factor
  of $0.5$ every fifty epochs of training. We used the Adam optimizer \cite{kingma2014adam} and a batch size of 64. 
  All training images were cropped and resized to a shape of $128 \times 128$ pixels, while a mesh of size $64 \times 64$
  was used in training. This allowed us to sample one keypoint for every four pixels in the UV space, making the mesh
  fairly high resolution. The mesh was initialized as a Gaussian surface, and was initially positioned so that it
  faces toward the camera. 

  \subsection{Datasets}
   We now note the face datasets that we used for our experiments. Certain among them contain side information, for instance multiple views of the same person, or videos of the same person. This side information was used for expression-identity disentanglement experiments, but not for the 3D lifting part. For the reconstruction results our algorithms were only provided with unstructured datasets, unless otherwise noted. 
   
   \begin{enumerate}
       \item \textbf{CelebA} \cite{liu2015faceattributes}: This dataset contains about 200,000 in-the-wild images,
         and is one of the datasets we use to train our DAE. A subset of this dataset, MAFL \cite{zhang2014eccv_mafl},
         was also released which contains annotations for five facial landmarks. We use the training set of MAFL
         in our evaluation experiments, and report results on the test set. Further, as MAFL is a subset of CelebA, 
         we removed the images in the MAFL test set from the CelebA training set before training the DAE.
       \item \textbf{Multi-PIE} \cite{multipie2010}: Multi-PIE contains images of 337 subjects of 7 facial expressions, each of which is captured under 15 viewpoints and 19 illumination conditions simultaneously.  
       \item \textbf{AFLW2000-3D} \cite{zhu2017face}: This dataset consists of 3D fitted faces for the first 2000 images of the AFLW dataset. In this paper, we employ it for evaluation of our learned shapes using 3D landmark localization errors. 
       
   \end{enumerate}
   
  \subsection{Qualitative Results}
  
   In this section, we show examples of the learned 3D shapes. 
   Figure \ref{fig:rotation} shows visualizations of
   reconstructed faces from various yaw angles using a model that was trained only on CelebA images. 
   We see that the model learns a shape that expresses the input well.
   However, when using no pose information from Multi-PIE, and the completely unsupervised nature of our alignment, 
   it is not able to properly decode side poses. This drawback is quickly overcome when we add weak pose
   supervision from the Multi-PIE dataset, as seen in Figure \ref{fig:refinement}.

   \begin{figure*}[ht!]
    \centering
    \includegraphics[width=.9\textwidth]{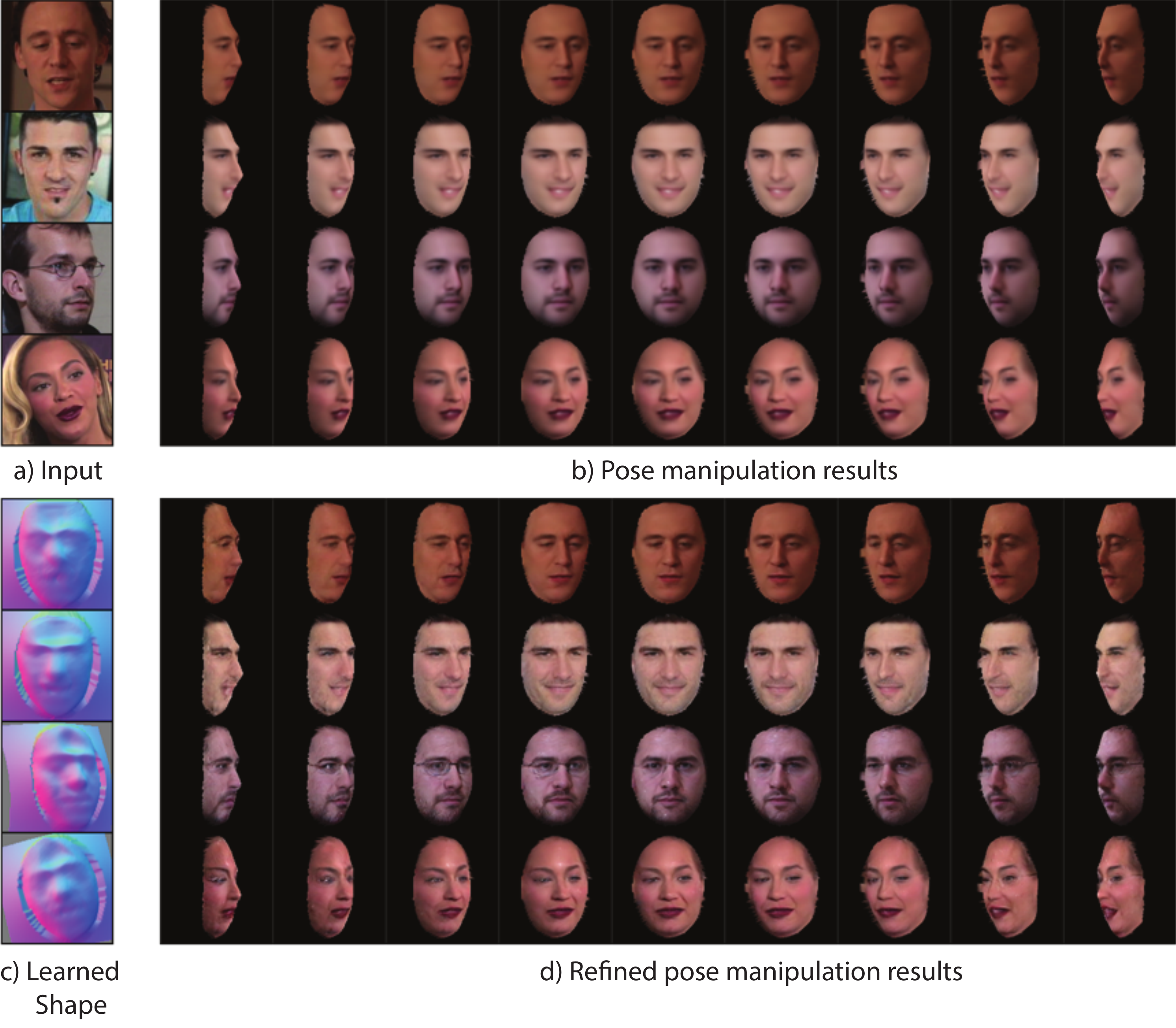}
    \captionof{figure}{Changing Pose with LAE. Given input face image (a), LAE learns to recover the 3D shape (c), with which we can manipulate the pose of the faces (b). With the additional refinement network, we can enhance the manipulated face image by adding facial details (d) that better preserve the characteristic features of the input faces. }
    \label{fig:results_pose}
  \end{figure*}
     
  \begin{figure*}[ht!]
    \centering
    \includegraphics[width=.85\textwidth]{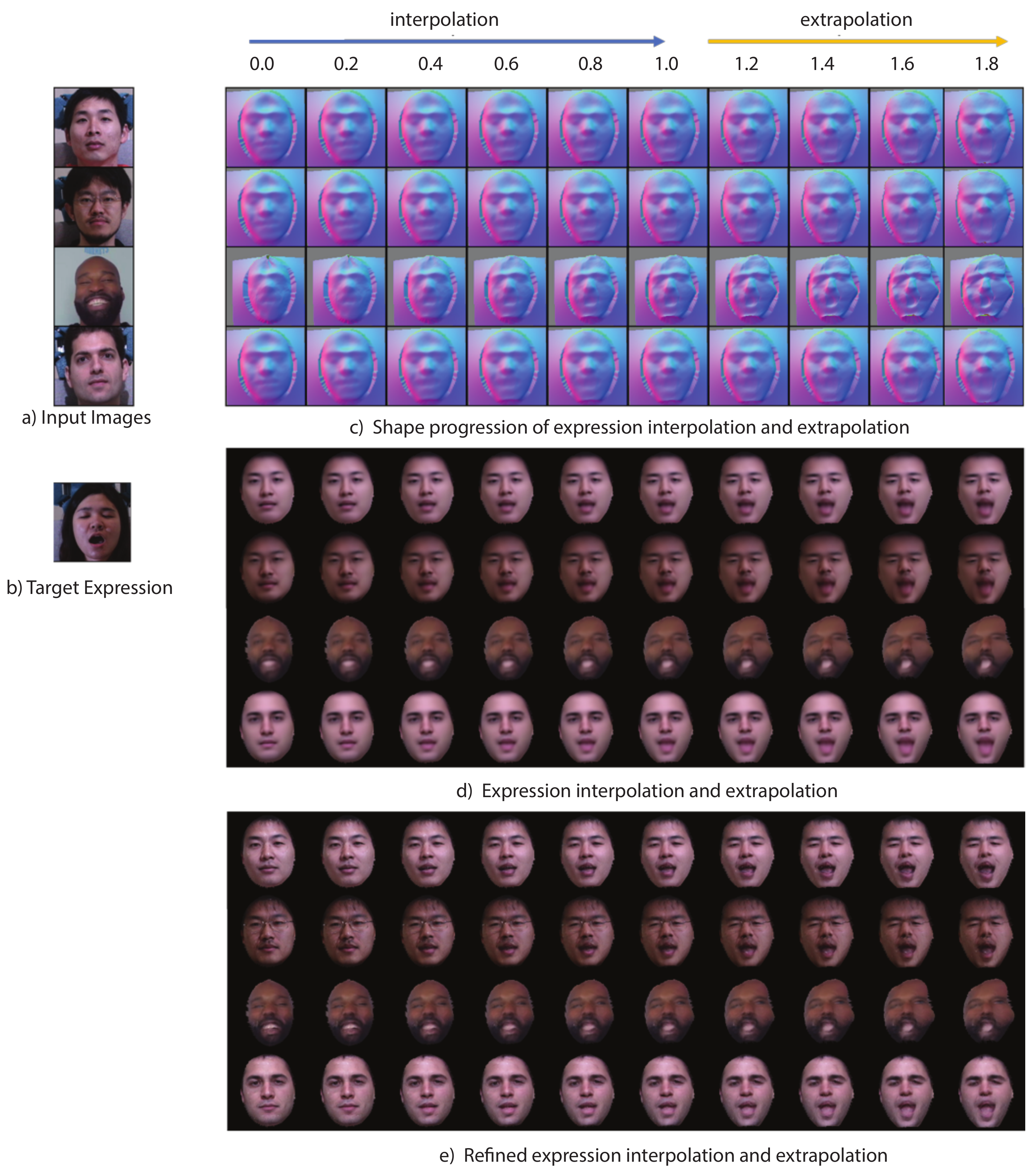}
    \captionof{figure}{Changing Expression with LAE. With LAE we can perform facial expression interpolation and extrapolation. Given the input faces (a), we can simply transfer the facial expression from another image (b) onto (a) with varying intensities by manipulating the learned expression representations. From (c,d,e) we observe continuous facial expression transformation from the input (a) to the target (b) (column 1 to 6), as well as realistic expression enhancements (column 7 to 10) via latent representation extrapolation (note the mouth and the eyes region).}
    \label{fig:results_expression}
  \end{figure*}
  
    \section{Face manipulation results}
  \label{sec:manip_results}
  
  In this section, we show some results of manipulating the expression and pose latent spaces. In Figure \ref{fig:results_pose} (b), we visualize the decoded 3D shape from input images in \ref{fig:results_pose} (a) from various camera angles. Furthermore, in Figure \ref{fig:results_pose} (d), we show results after passing the visualizations in Figure \ref{fig:results_pose} (b) through the refinement network. 
  
\begin{figure*}[ht!]
    \centering
    \includegraphics[width=.85\textwidth]{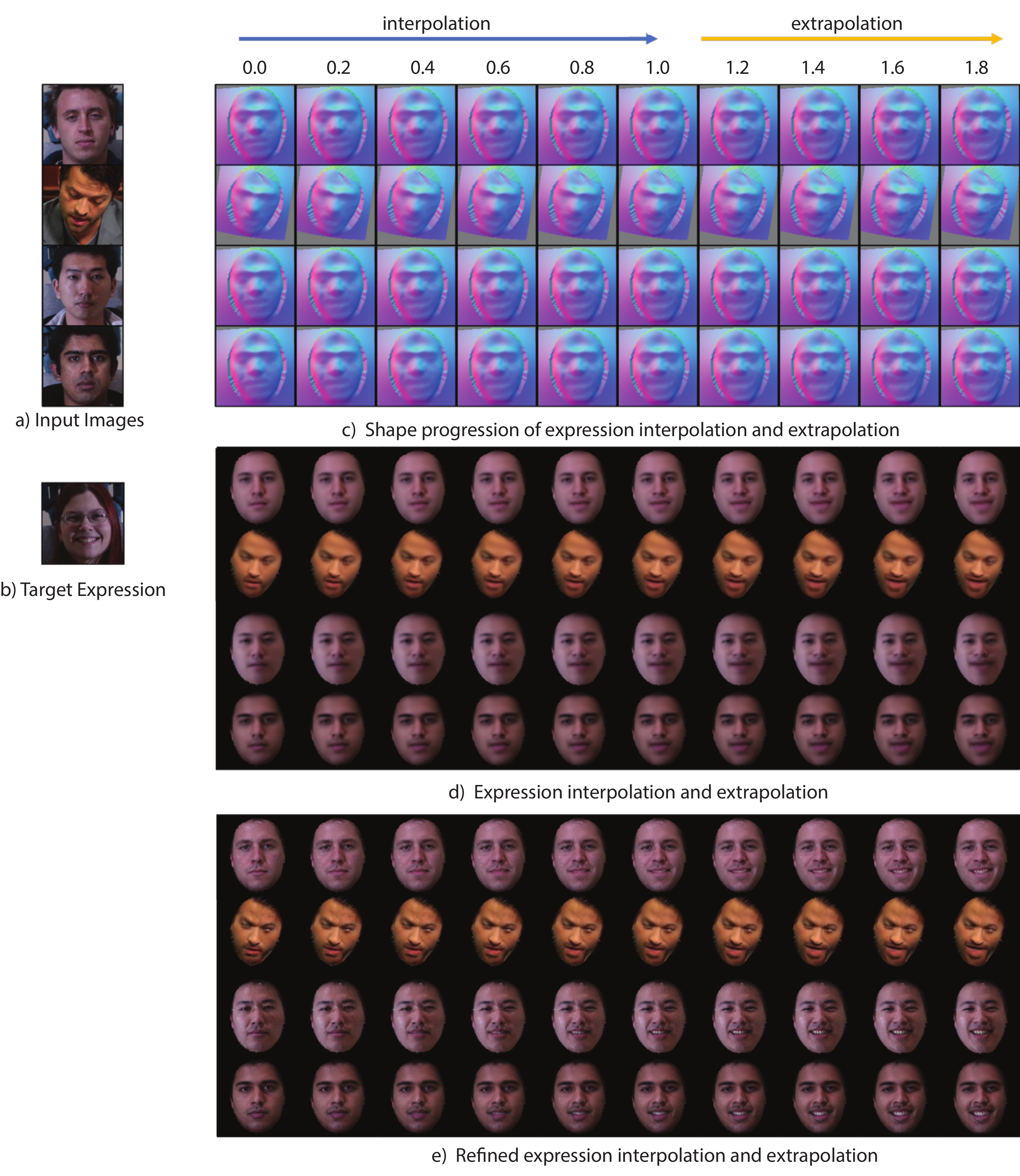}
    \captionof{figure}{Changing Expression with LAE. With LAE we can perform facial expression interpolation and extrapolation. Given the input faces (a), we can simply transfer the facial expression from another image (b) onto (a) with varying intensities by manipulating the learned expression representations. From (c,d,e) we observe continuous facial expression transformation from the input (a) to the target (b) (column 1 to 6), as well as realistic expression enhancements (column 7 to 10) via latent representation extrapolation (note the mouth and the eyes region).}
    \label{fig:results_expression2}
  \end{figure*}

  Similarly, in Figures \ref{fig:results_expression} and \ref{fig:results_expression2} (a)-(e), we interpolate over the expression latent space from each of the images in (a) to the image in (b), and visualize the shape at each intermediate step in Figure (c), the output in (d), and the refined output in (e). 
  
  
  Finally, in Figure \ref{fig:all_interpolation}, we interpolate over all three latent spaces---texture, pose, and shape.

   \begin{table}[ht!]
    \centering
    \begin{tabular}{l r}
     \toprule
        Method & NME  \\
     \midrule
        Thewlis \etal (2017) \cite{thewlis2017iccv} & $6.67$ \\
        Thewlis \etal (2018) \cite{thewlis2017nips} & $5.83$ \\
        Jakub \etal (2018) \cite{jakub2018nips} & $2.54$ \\  
        \addlinespace[4pt]        
        Shu \etal (2018), DAE, no regressor \cite{shu2018eccv} & $7.54$ \\
        Shu \etal (2018), DAE, with regressor \cite{shu2018eccv} & $5.45$ \\
        \addlinespace[4pt]
        LAE, CelebA (no regressor) & $7.96$ \\
        LAE, CelebA (with regressor) & $6.01$ \\
     \bottomrule
    \end{tabular}
    \caption{2D landmark localization results for the proposed LAEs compared with other state-of-the-art
      approaches. All numbers signify the average error per landmark normalized by the inter-ocular distance, 
      over the entire dataset.}
    \label{tab:mafl_results}
   \end{table}

   \subsection{Landmark Localization}
    Our system allows us to roughly estimate landmarks, by annotating them only once in the aligned, canonical space,
    as also shown by \cite{shu2018eccv}. Here we further visualize detected landmarks using the learned 3D shape
    in Figure \ref{fig:aflw_landmarks} on some images from the AFLW2000-3D dataset.

  \subsection{Albedo-shading disentanglement}

\begin{figure*}[ht]
     \centering
     \includegraphics[width=0.9\linewidth]{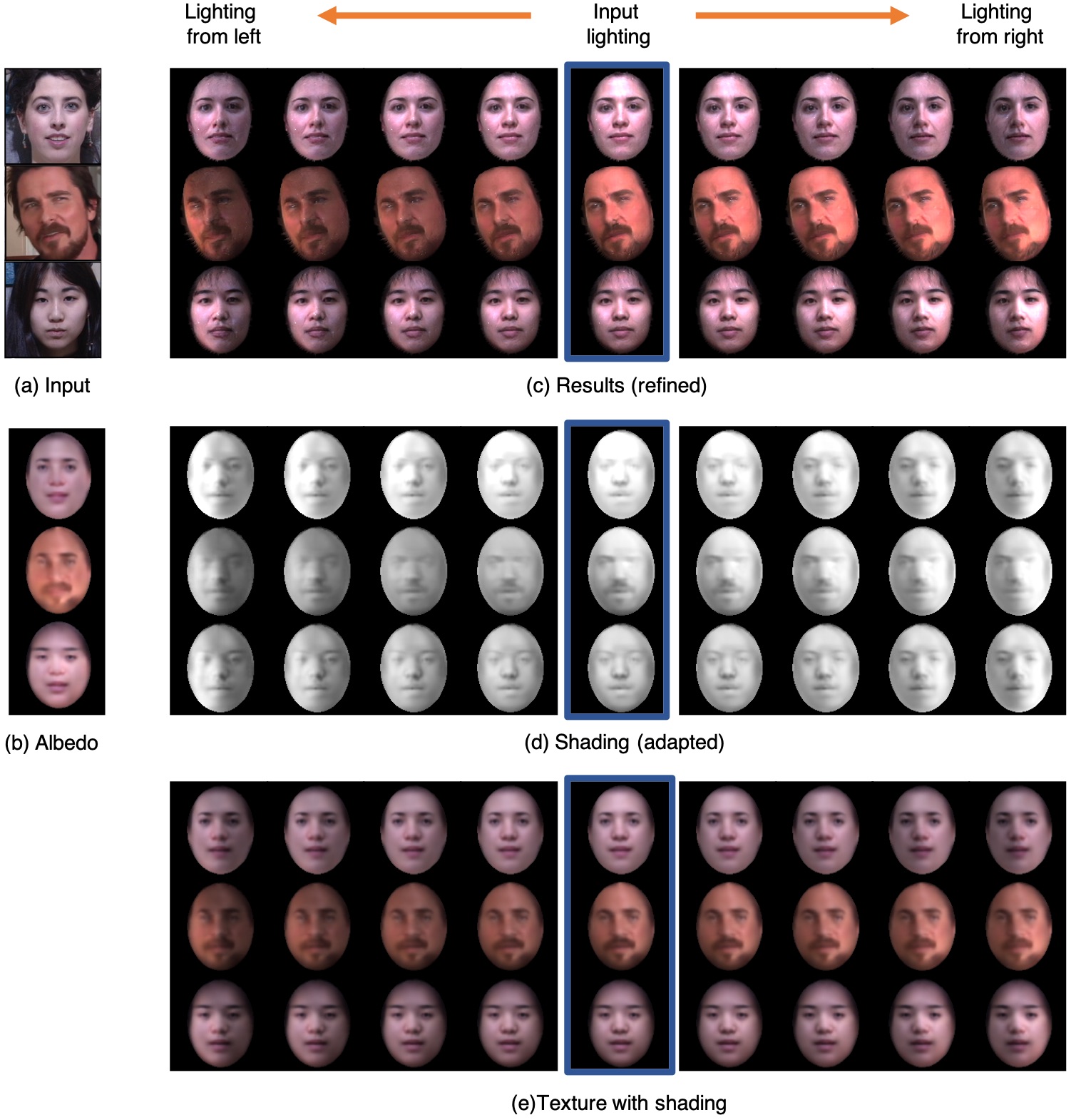}
     \caption{Lighting manipulation with LAE-lux. With disentangled albedo and shading and explicit shading representation using Spherical Harmonics, we can manipulate the illumination of faces. We show illumination editing of 3 examples from given input faces (a), to a hallucinated lighting from left ((c) - left side) and a hallucinated lighting from right ((c) - right side). Interpolation of Spherical Harmonics coefficients generates smooth transition of shading effect (d), combining with the learned albedo (b), we obtain the dense aligned texture with different illumination effect (e). Final results (c) are obtained by applying deformation learned in LAE to (e), and a refinement step. }
     \label{fig:illuminationchange}
\end{figure*}

    In Fig.~\ref{fig:illuminationchange} we show that with the disentangled physical representation for illumination, we can hallucinate illumination manipulation with LAE-lux.
 
  \subsection{Quantitative Analysis: Landmark Localization}
  
   We evaluate our approach quantitatively in terms of landmark localization. Specifically, we evaluate 
   on two datasets---the MAFL test set for 2D landmarks, and the AFLW2000-3D for 3D shape. In both cases, 
   as we do not train with ground-truth landmarks, we manually annotate, only once, the necessary landmarks 
   on the base shape as linear combinations of one or more mesh vertices. That is to say, each landmark 
   location corresponds to a linear combination of the locations of several vertices.
  
   We use five landmarks for the MAFL test set, namely the two eyes, the tip of the nose, and the ends of
   the mouth. Similarly to \cite{thewlis2017iccv, thewlis2017nips, shu2018eccv}, we evaluate the extent to which 
   landmarks are captured by our 3D shape model by training a linear regressor to predict them given
   the locations of the mesh vertices in 3D.

   We observe from Table \ref{tab:mafl_results} that our system is able to perform at-par with the DAE, which is our 
   starting model - and as such serves as the upper bound on the performance that we can attain. This shows that while being  able to successfully perform the lifting operation, we do not sacrifice localization accuracy. The small increase in error can  be attributed to the fact that perfect 
   reconstruction of a system is nearly impossible with a low-dimensional shape model. Furthermore we use a feedforward, single-shot camera and shape regression network, while in principle this is a problem that could require iterative model fitting techniques to align a 3D deformable model to 2D landmarks \cite{pavlakos2017object3d}. 
   
   We  report localization results in 3D on 21 landmarks that feature in the AFLW2000-3D dataset. 
   As our unsupervised system is often unable to locate human ears, the learned face model
   does not account for them in the UV space. This makes it impossible to evaluate landmark localization for
   points that lie on or near the ears, which is the case for two of these landmarks. Hence, for the 
   AFLW2000-3D dataset, we report localization accuracies only for 19 landmarks. 
   Furthermore, as an evaluation of the discovered shape, we also show landmark localization results 
   after rigid alignment (without reflection) of the predicted landmarks with the ground truth. 
   We perform Procrustes analysis, with and without adding rotation to the alignment, the latter giving 
   us an evaluation of the accuracy of pose estimation as well. 
   
   \mycomment{
   \begin{figure*}[ht!]
      \centering
      \includegraphics[width=\linewidth]{figures/ExpressionsNew2.png}
       \caption{Expression manipulation: given an observed image, we lift it in 3D and manipulate it in the learned basis for expressions. Here we add to the reconstructed surface a component lying in the direction of the `open mouth' expression, and then render an image based on the new geometry. This allows us to synthesize photorealistic images that are geometrically controlled.
       }
       \label{fig:expressions}
   \end{figure*}
   }

   Table \ref{tab:aflw3d_results} also demonstrates the gain achieved by adding weak supervision via the Multi-PIE dataset. We see that the mean NMEs for LAEs trained with and without the Multi-PIE dataset increase as the yaw angle increases. This is also visible in our qualitative results shown in Fig.~\ref{fig:refinement}, where we visualize 
   the discovered shapes for both of these cases. 
   
\begin{table*}[ht!]
    \centering
    \begin{tabular}{
        l 
        l 
        S
        S
        S
        S
      }
      \toprule
      \multirow{2}{*}{\textbf{Method}} &
      \multirow{2}{*}{\textbf{Rotation}} & 
      \multicolumn{4}{c}{Yaw angle}  \\\cmidrule(l{7pt}){3-6}
      & & 
      \multicolumn{1}{r}{$[0, 30]$} &
      \multicolumn{1}{r}{$(30, 60]$} & 
      \multicolumn{1}{r}{$(60, 90]$} & 
      \multicolumn{1}{r}{All} \\ 
      \midrule
      \multirow{2}{*}{\shortstack[l]{3DDFA \cite{zhu2017face}\\(supervised)}} & Y & 4.25 \pm 0.95 & 4.34 \pm 1.04 & 4.39 \pm 1.35 & 4.28 \pm 1.03 \\
              & N & 12.51 \pm 6.40 & 23.20 \pm 5.92 & 32.55 \pm 3.85 & 17.31 \pm 9.30 \\
      \addlinespace[4pt]              
      \multirow{2}{*}{\shortstack[l]{PRNet \cite{feng2018prn}\\(supervised)}} & Y & 4.88 \pm 1.24 & 6.94 \pm 2.83 & 10.51 \pm 5.31 & 6.01 \pm 3.08 \\
              & N & 7.17 \pm 3.45 & 10.96 \pm 5.00 & 16.34 \pm 8.91 & 9.11 \pm 5.66 \\
      \addlinespace[4pt]              
      \multirow{2}{*}{\shortstack[l]{3D-FAN \cite{bulat2017far}\\(supervised)}} & Y & 2.73 \pm 1.38 & 2.48 \pm 2.24 & 3.74 \pm 2.95 & 2.84 \pm 1.92 \\
              & N & 7.51 \pm 2.21 & 7.06 \pm 3.94 & 8.75 \pm 4.53 & 7.61 \pm 3.10 \\
      \midrule
      \multirow{2}{*}{LAE (64) CelebA} & Y & 6.86 \pm 1.07 & 9.01 \pm 1.07 & 10.91 \pm 1.37 & 7.89 \pm 1.89 \\
              & N & 9.29 \pm 4.90 & 20.98 \pm 7.74 & 37.62 \pm 7.50 & 15.85 \pm 11.89 \\
      \addlinespace[4pt]        
      \multirow{2}{*}{LAE (128) CelebA} & Y & 6.02 \pm 1.04 & 7.91 \pm 1.04 & 9.58 \pm 1.32 & 6.92 \pm 1.73 \\
              & N & 8.41 \pm 4.96 & 19.56 \pm 7.97 & 36.31 \pm 7.78 & 14.80 \pm 11.80 \\
      \addlinespace[4pt]        
      \multirow{2}{*}{LAE (128) MultiPIE} & Y & 6.85 \pm 0.85 & 7.94 \pm 0.97 & 9.02 \pm 1.26 & 7.39 \pm 1.25 \\
              & N & 9.80 \pm 4.88 & 13.87 \pm 6.51 & 24.19 \pm 8.72 & 12.78 \pm 7.83 \\   
      \addlinespace[4pt]        
      \multirow{2}{*}{\shortstack[l]{LAE (128)\\CelebA+MultiPIE}} & Y & 6.83 \pm 0.96 & 8.41 \pm 1.15 & 9.83 \pm 1.65 & 7.59 \pm 1.60 \\
              & N & 9.11 \pm 4.54 & 13.60 \pm 6.08 & 24.62 \pm 8.37 & 12.33 \pm 7.84 \\                
      \bottomrule
    \end{tabular}
    \caption{Mean 3D landmark localization errors, after Procrustes analysis, normalized by bounding box size and averaged over the entire AFLW2000-3D test set. The number in brackets for the LAEs refers to the dimension of the latent space for the rigid and non-rigid components of the deformable model. The second column specifies whether rotation is included
    in the Procrustes analysis. We also note the training dataset used for training each LAE.}
    \label{tab:aflw3d_results}
\end{table*}

\section{Conclusion}
In this work we have introduced an unsupervised method for lifting an object category into a 3D representation, allowing us to learn a 3D morphable model of faces from an unorganized photo collection. We have shown that we can use the resulting model for controllable manipulation and editing of observed images.

Deep image-based generative models have shown the ability to deliver photorealistic synthetsis results with substantially  more diverse categories than faces \cite{brock2018large,karras2018iclr} - we anticipate that their combination with  3D representations like LAEs will further unleash their potential for controllable image synthesis.

\appendix

\section{Additional Details}
 In this section, we note some additional implementation details. 
 \subsection{Data Processing}
  \label{sec:data}
  In our experiments, we used images of size $128 \times 128 \times 3$ pixels, which were cropped from the CelebA and MultiPIE datasets using ground-truth bounding boxes. 
  
  For CelebA images, the cropping was performed by extracting a square patch around the face with side-length equal to the length of the longer side of the bounding box. It was then adjusted so that it lies entirely inside the image (by translating it horizontally or vertically, or even scaling it down if necessary). Finally, we tightened the resulting box by $12$ pixels from each side as the bounding boxes are quite loose crops, and resized the resulting square image to $128 \times 128$.  We use all images from CelebA for training (about $200,000$ images) except the MAFL test set which is contained entirely in CelebA ($1000$ images). 
  
  For MultiPIE dataset, we crop the face images according to landmarks positions on the eyes, the corner of mouth, and the width of the frontal face. Specifically, we use the mean coordinates of the 4 landmarks as the center of the crop, and use $1.4 \times$ the width of the face as the width of the images. We use the method proposed in \cite{bulat2017far} to detect the landmarks. For each person, the crop is identical across all illumination condition for the same camera.

 \subsection{Architecture Details} 
  \label{sec:arch}
  We used convolutional encoders and decoders similar to the ones described in \cite{shu2018eccv}. We detail the architectures here again for completeness. The convolutional encoder architecture is---
  \begin{verbatim}
Conv(32)-LeakyReLU-Conv(64)->
  ->BN-LeakyReLU-Conv(128)->
  ->BN-LeakyReLU-Conv(256)->
  ->BN-LeakyReLU-Conv(256)->
  ->BN-LeakyReLU-Conv(Nz)->
  ->Sigmoid;
  \end{verbatim}
  while the convolutional decoder architecture is---
\begin{verbatim}
ConvT(512)-BN-ReLU-ConvT(256)->
  ->BN-ReLU-ConvT(128)->
  ->BN-ReLU-ConvT(64)->
  ->BN-ReLU-ConvT(32)->
  ->BN-ReLU-ConvT(32)->
  ->BN-ReLU-ConvT(Nc)->
  ->Threshold(0,1).
\end{verbatim}

 \subsection{Refinement Networks}
  The refinement set-up consists of a \emph{generator} network, and a \emph{discriminator} network. The generator is a standard UNet\cite{ronneberger2015u} for $128 \times 128$ images that  are downsampled to $1 \times 1$ in the innermost latent layer. 
  
  The discrminiator is a PatchGAN discrminator\cite{isola2017image} with the following architecture---
  \begin{verbatim}
Conv(64)-LeakyReLU-Conv(128)-BN->
  ->LeakyReLU-Conv(256)-BN->
  ->LeakyReLU-Conv(512)-BN->
  ->LeakyReLU->Conv(1)
  \end{verbatim}
  
  In all these descriptions, \texttt{Conv(x)} signifies a 2D convolution layer with \texttt{x} channels, a kernel size of $4 \times 4$, a stride of $2$, and a padding of $1$. Similarly for \texttt{ConvT(x)}, except that it signifies a deconv layer.

 \subsection{Implementation Details}
  \label{sec:implementation}
  We implemented our system in Python 3.6 using the PyTorch library. We use convolutional, activation, and batch norm layers predefined in the \texttt{torch.nn} module, and take advantage of the Autograd\cite{paszke2017automatic} framework to take care of the gradients required by backpropagation.
  
  \subsection{Rotation Modeling}
   Modelling rotations using quaternions has several advantages over modelling them using Euler angles, including computational ease, less ambiguity, and compact representation\cite{Dam98quaternions}. Quaternions were also employed by \cite{cmrKanazawa18} to model mesh rotations. 
   Following these works, we also use quaternions in our framework to model rotations, by regressing them from the camera latent space, and normalizing them to unit length. 
   
  \subsection{The Neural Mesh Renderer}
   The Neural Mesh Renderer\cite{kato2018renderer} is a recently proposed module that can be inserted into a neural network to enable end-to-end training with a rendering operation. The renderer proposes approximate gradients to learn texture and shape given the output rendering. The original module was released in Chainer\cite{kato2018github}, but we use a PyTorch port of this module, which is a publicly-available re-implementation\cite{nr2018pytorch}. The renderer in our framework accepts a texture image, the mean shape, the deviation from the mean shape, and the camera parameters to output a 2D reconstruction of the original image.

 \subsection{Training Procedure}
  \label{sec:training}
  To train the LAE, we first train a DAE on the training data. We then fix the DAE and use it to extract dense correspondences between the image space and the canonical space. These correspondences are used in the objective of the 3D reprojection loss (Equations 6 and 7 in the paper). 
  
  To obtain image-specific camera, translation, and shape estimates, we train another convolutional encoder. This encoder learns a disentangled latent space where the shape estimates and camera and translation estimates are encoded by different vectors. For the MultiPIE experiments, the shape latent vector is further divided into identity and expression vectors. We use linear layers to regress camera, translation, and shape estimates from their latent encodings. 
  
  We train our system using the Adam\cite{kingma2014adam} optimizer for all learnable parameters. We start with a learning rate of $0.0001$, which is decayed every $50$ training epochs by a factor of $0.5$. We train for a total of $400$ epochs.

\bibliographystyle{ieee}
\bibliography{bibliography}

\end{document}